\documentclass[letterpaper]{article} % DO NOT CHANGE THIS

\usepackage[]{aaai24}  % DO NOT CHANGE THIS
\usepackage{times}  % DO NOT CHANGE THIS
\usepackage{helvet}  % DO NOT CHANGE THIS
\usepackage{courier}  % DO NOT CHANGE THIS
\usepackage[hyphens]{url}  % DO NOT CHANGE THIS
\usepackage{graphicx} % DO NOT CHANGE THIS
\usepackage{comment}
\usepackage{adjustbox}

\usepackage{amsmath}
\usepackage{natbib}  % DO NOT CHANGE THIS AND DO NOT ADD ANY OPTIONS TO IT
\usepackage{caption} % DO NOT CHANGE THIS AND DO NOT ADD ANY OPTIONS TO IT
\usepackage{algorithm}
\usepackage{algorithmic}
\usepackage{amsmath}
\usepackage{amssymb}
\usepackage{dsfont}
\usepackage{array}
\usepackage{xcolor}
\usepackage{newfloat}
\usepackage{listings}
\DeclareCaptionStyle{ruled}{labelfont=normalfont,labelsep=colon,strut=off} % DO NOT CHANGE THIS
\floatstyle{ruled}
\newfloat{listing}{tb}{lst}{}
\floatname{listing}{Listing}
\title{Understanding (Un)Intended Memorization in Text-to-Image Generative Models}
\author{
    Ali Naseh,\textsuperscript{\rm 1}
    Jaechul Roh, \textsuperscript{\rm 1}
    Amir Houmansadr \textsuperscript{\rm 1}
}
\affiliations{
    \textsuperscript{\rm 1} University of Massachusetts Amherst\\
    anaseh@cs.umass.edu, jroh@umass.edu, amir@cs.umass.edu
}

\makeatletter% <-------------------
    \renewcommand{\copyright@on}{F}
\makeatother
\begin{document}

\maketitle

\section{Abstract}

Multimodal machine learning, especially text-to-image models like Stable Diffusion and DALL-E 3, has gained significance for transforming text into detailed images.
 Despite their growing use and remarkable generative capabilities, there is a pressing need for a detailed examination of these models' behavior, particularly with respect to memorization. Historically, memorization in machine learning has been context-dependent, with diverse definitions emerging from classification tasks to complex models like Large Language Models (LLMs) and Diffusion models. Yet, a definitive concept of memorization that aligns with the intricacies of text-to-image synthesis remains elusive. This understanding is vital as memorization poses privacy risks yet is essential for meeting user expectations, especially when generating representations of underrepresented entities. In this paper, we introduce a specialized definition of memorization tailored to text-to-image models, categorizing it into three distinct types according to user expectations. We closely examine the subtle distinctions between intended and unintended memorization, emphasizing the importance of balancing user privacy with the generative quality of the model outputs. Using the Stable Diffusion model, we offer examples to validate our memorization definitions and clarify their application.

\section{Introduction}

The landscape of machine learning is currently witnessing an explosion of interest in multimodal systems, with text-to-image models such as the Stable Diffusion~\cite{esser2021taming}, Midjourney~\cite{midjourney2022} and DALL-E 3~\cite{shi2020improving} at the forefront. These models exhibit remarkable capabilities, transforming textual descriptions into vivid images with astonishing accuracy and creativity. As these systems become increasingly prevalent, their widespread adoption across various domains necessitates a thorough examination of their behavior. One critical aspect that has received significant attention is the concept of memorization. This phenomenon, while crucial in shaping the models' performance, presents complex challenges and opportunities that merit a deeper investigation.

\begin{table}[ht]
    \centering
    \caption{Definitions for different types of Memorization in Text-to-Image Models}
    \label{tab:notations_terms}
    \begin{tabular}{>{\centering\arraybackslash}m{0.4\columnwidth} >{\centering\arraybackslash}m{0.5\columnwidth}}
    \hline
    \textbf{Type} & \textbf{Definition} \\ \hline
    \textbf{Explicit Intended Memorization}&  Memorization of image features presented in user prompts, expected to be deliberately memorized by the model, often involves well-known entities. \\ \hline
    \textbf{Implicit Intended Memorization} & Involves the model's indirect memorization of image features anticipated by users but not explicitly stated in the prompt. \\ \hline
    \textbf{Unintended Memorization} & Involves features the model has inadvertently memorized, which are neither expected by users nor included in the input prompt or related concepts. \\
    \hline
    \end{tabular}
\end{table}
\begin{table*}[ht]
\centering
\caption{Images produced by the pre-trained Stable Diffusion, along with the textual features (\(F_T\)), image features (\(F_I\)) and the topic words extracted from the input prompt using GPT-4.}
    \label{tab:gpt4_example}
    \begin{adjustbox}{width=\textwidth,center}
    \begin{tabular}{>{\centering\arraybackslash}m{0.5\columnwidth} >{\arraybackslash}m{1.3\columnwidth}}
    \hline
    \textbf{Input Prompt} & \small \texttt{Winter season in the city during Christmas times with jazz and music} \\ \hline
    \textbf{Generated Image} & {\includegraphics[width=9cm]{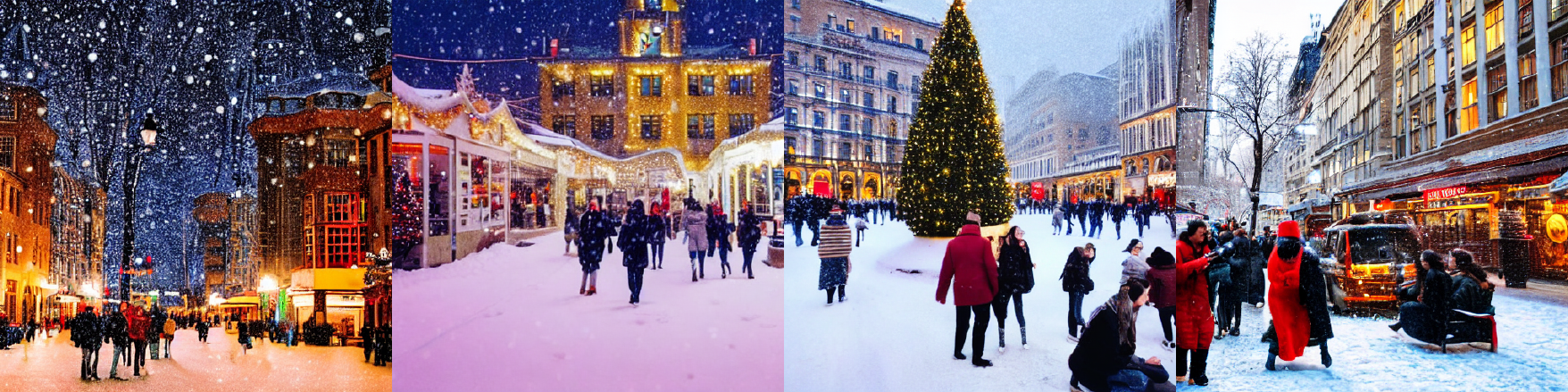}} \\ \hline
    \textbf{Extracted Topics/Concepts} & \small \texttt{\{Winter Season and Cityscapes\}, \{Christmas Celebrations in Urban Settings\}, \{Jazz and Music During Christmas\}, \{Cultural Aspects of Christmas\}, \{Seasonal Events and Festivals\},
    \{Seasonal Events and Festivals\},
    \{Impact of Music on Holiday Atmosphere\},
    \{Winter Activities and Entertainment in Cities\}
    }
    \\ \hline
    \(F_T\) & \small \texttt{\{Christmas Times\}, \{Inter Season, City\}, \{Jazz and Music\}, \{Imagery and Atmosphere\}, \{Temporal Setting\}, \{Cultural and Festive Elements\}, \{Sensory Details\}, \{Contextual Ambiguity\}} \\ \hline
    \(F_I\) & \small \texttt{\{Snowfall\}, 
    \{Snow\}, 
    \{Urban Environment\}, 
    \{People\}, 
    \{Christmas Decorations\}, 
    \{Evening or Nighttime Setting\}, 
    \{Winter Clothing\}, 
    \{Street Lights\}, 
    \{Christmas Market or Stalls\}, 
    \{Trees\}, 
    \{Cafés and Shops\}
    } \\ 
    \hline
    \end{tabular}
    \end{adjustbox}
\end{table*}

In the realm of machine learning, memorization is a concept that has been extensively studied across various tasks, from basic classification~\cite{feldman2020does, carlini2019secret, feldman2020neural} to sophisticated models like LLMs~\cite{carlini2021extracting, mireshghallah2022empirical} and Diffusion models~\cite{carlini2023extracting}. Each task has brought forth its own set of definitions for memorization, reflecting the particular needs and challenges of the task at hand. However, in the context of text-to-image models, a comprehensive definition of memorization is still missing. The rapid progress in this area, with its complex generative processes, demands more attention from researchers. There is a clear need for a definition of memorization that fully encompasses the complexity of these text-to-image models.

Memorization within machine learning models presents a double-edged sword; on one side, it poses risks for individual privacy~\cite{shokri2017membership} and the potential leakage of sensitive information~\cite{carlini2021extracting}. On the other hand, efforts to mitigate memorization, including deduplication techniques~\cite{kandpal2022deduplicating}, can inadvertently impair the model's performance~\cite{carlini2023extracting}, particularly in accurately rendering images of entities that are not well-represented in the training data. \citeauthor{feldman2020does}~\shortcite{feldman2020does}  demonstrated that in classification tasks, particularly in extreme multi-label classification, for subclasses with limited data (often referred to as the "long tail" of data distribution), the model might be unable to make accurate predictions unless it has encountered at least one sample from that subclass. The same principle applies to text-to-image generative models. For outliers in the training data that appear infrequently, users expect the model to memorize these instances and produce near-identical images.

In certain scenarios, user expectations align with the model's ability to reproduce specific requested features in the prompt, necessitating a certain degree of memorization. Text-to-image models are expected not only to faithfully generate what is explicitly mentioned in the prompt but also to exhibit creativity in conjuring up accompanying features~\cite{li2022stylet2i}. This multifaceted aspect emphasizes the importance of differentiating between intended and unintended memorization. For a mitigation strategy to be effective without compromising the generative quality of these models, it must be founded on a comprehensive understanding of memorization that encompasses the full spectrum of scenarios and user expectations.

In this paper, we present a general definition of memorization specifically designed for text-to-image models, and categorize memorization into three distinct types according to user expectations and various operational scenarios. We further explore these categories using the Stable Diffusion model—one of the most popular and accessible open-source text-to-image models—to provide illustrative examples that enhance the reader's understanding of our proposed definitions.

\begin{table}[ht]
    \centering
    \caption{Notations and Corresponding Terms}
    \label{tab:notations_terms}
    \begin{tabular}{>{\centering\arraybackslash}m{0.2\columnwidth} >{\centering\arraybackslash}m{0.7\columnwidth}}
    \hline
    \textbf{Notation} & \textbf{Term} \\ \hline
    \( M_{E} \) & Explicit Intended Memorization \\
    \( M_{I} \) & Implicit Intended Memorization \\
    \( M_{U} \) & Unintended Memorization \\
    \( F_{I} \) & Union of the image features of all training images \\
    \( F_{T} \) & Text features of the prompt\\
    \( F'_{I} \) & Union of the image features of all training images (duplicates removed) \\
    \( f_i \) & A single image feature \\
    \( f_t \) & A single text feature \\
    \hline
    \end{tabular}
\end{table}
\section{Related Works}

\subsection{Text-to-Image Synthesis}
Early research~\cite{nichol2021glide, ramesh2022hierarchical, gu2022vector, rombach2022high} highlights significant progress in text-to-image models. GLIDE \cite{nichol2021glide} initiates the use of diffusion models in text-to-image tasks. Imagen~\cite{saharia2022photorealistic} build on GLIDE's~\cite{nichol2021glide} framework, incorporating frozen language model to lessen computational load. DALLE-2 \cite{ramesh2022hierarchical} introduces a two-step process, first converting text to image embeddings, then using these for image generation via diffusion models. Stable Diffusion~\cite{rombach2022high} tackled the issue of low-resolution outputs by training diffusion models in latent space, employing an autoencoder-diffusion-autoencoder structure for enhanced image quality. 

% Early studies \cite{nichol2021glide, ramesh2022hierarchical, gu2022vector, rombach2022high} demonstrate marked advancements in text-to-image tasks. Initially, GLIDE \cite{nichol2021glide} stands out as a pioneering model, introducing an innovative approach that employs diffusion models. Imagen \cite{saharia2022photorealistic} introduces a modification to GLIDE's \cite{nichol2021glide} text encoder framework. It employs frozen large language models \cite{zhang2023text}, a strategic shift that results in a considerable reduction in computational demands.

% DALLE-2 \cite{ramesh2022hierarchical} introduces a sophisticated 2-stage model. The initial phase involves the transformation of text into CLIP \cite{radford2021learning} image embedding. In the decoding phase, the model intakes the embedding as a conditional input, leveraging the diffusion model for image generation. 

% Addressing the prevalent challenge of producing low-resolution images from prior studies \cite{nichol2021glide, saharia2022photorealistic, ramesh2022hierarchical}, the Stable Diffusion framework \cite{rombach2022high} adeptly employs VQ-GAN \cite{esser2021taming} to facilitate the training of the diffusion model in the latent space. This intricate architecture positions an autoencoder on either side of the diffusion model, thereby enabling the diffusion model to train the latent embedding. 

\subsection{Memorization in Text-to-Image Synthesis}
Recent studies~\cite{carlini2023extracting, somepalli2023diffusion, somepalli2023understanding} identify memorization in text-to-image models, largely attributed to duplications in the LAION dataset~\cite{schuhmann2021laion, schuhmann2022laion}. \citeauthor{carlini2023extracting}~\shortcite{carlini2023extracting} highlight that diffusion models, while generating 500 variations per caption using distinct seeds, can reproduce exact training images. When 10 or more generations are near-identical, they are classified as memorized. \citeauthor{somepalli2023diffusion}~\shortcite{somepalli2023diffusion} presents case studies on memorization in Stable Diffusion~\cite{rombach2022high}, comparing generated images to their nearest counterparts from the training set via SSCD representation~\cite{pizzi2022self}. The matched images often correspond to different captions~\cite{somepalli2023diffusion}. Furthermore, \citeauthor{somepalli2023understanding}~\shortcite{somepalli2023understanding} underscores Stable Diffusion \cite{rombach2022high} memorization tendencies across caption variations.

\subsection{Evaluation Metrics}
Metric to measure memorization in text-to-image models varies across different works. \citeauthor{carlini2023extracting}~\shortcite{carlini2023extracting}, for instance, adopt the Euclidean 2-norm distance as a measure to determine the similarity between two images, a methodology inspired by \citeauthor{balle2022reconstructing}~\shortcite{balle2022reconstructing}. Conversely, \citeauthor{somepalli2023diffusion}~\shortcite{somepalli2023diffusion} propose the utilization of the dot product of SSCD representations~\cite{pizzi2022self} as a mechanism to discern duplications. Building upon their research, \citeauthor{somepalli2023understanding}~\shortcite{somepalli2023understanding} leverage the Frechet Inception Distance (FID)~\cite{heusel2017gans} score as a benchmark to assess the fidelity of synthesized images. It is noteworthy that a lower FID score is indicative of a superior image quality and diversity. \citeauthor{zhang2023forget}~\shortcite{zhang2023forget} introduces the notion of a ‘memorization score’ to assess how much content synthesized images retain, by comparing the cosine similarity of concept embeddings \textemdash~from the original and the sanitized ‘forgotten’ images \textemdash ~to those of the training dataset embeddings.  

\begin{comment}
\subsection{GPT-4 Feature Extraction}

In the recent release of the multi-modal GPT-4 model \cite{chatgpt2023}, there has been a significant improvement in its abilities to caption text and images \cite{liu2023summary}. To assess its capabilities, we employ the model to describe various aspects of the image displayed in \autoref{tab:gpt4_example}, the attributes of the input prompt, and the broader themes associated with the input prompt. GPT-4 \cite{chatgpt2023} generates comprehensive descriptions, recognizing elements such as "Christmas Times" and "Imagery and Atmosphere" in text, as well as "Urban Environment", "Christmas Decorations", and "Street Lights" in images. Furthermore, the model offers thorough analysis of each individual feature along with a syntheized summary of all attributes. Additionally, the model is proficient at extracting relevant topics and concepts from the provided input prompt, which greatly assists in determining whether the generated images aligns especially with implicit intended memorization or not. 

\end{comment}

\begin{table*}[ht]
\centering
\caption{Example of Explicit Intended Memorization \(M_E\) based on the images created using pre-trained Stable Diffusion.}
    \label{tab:M_E_orig}
    \begin{adjustbox}{width=\textwidth,center}
    \begin{tabular}{>{\centering\arraybackslash}m{0.5\columnwidth} >{\arraybackslash}m{1.3\columnwidth}}
    \hline
    \textbf{Memorization Type} & Explicit Intended Memorization (\(M_E\)) \\ \hline
    \textbf{Input Prompt} & \small \small \texttt{High quality image of Elon Musk} \\ \hline
    \textbf{Generation} & {\includegraphics[width=9cm]{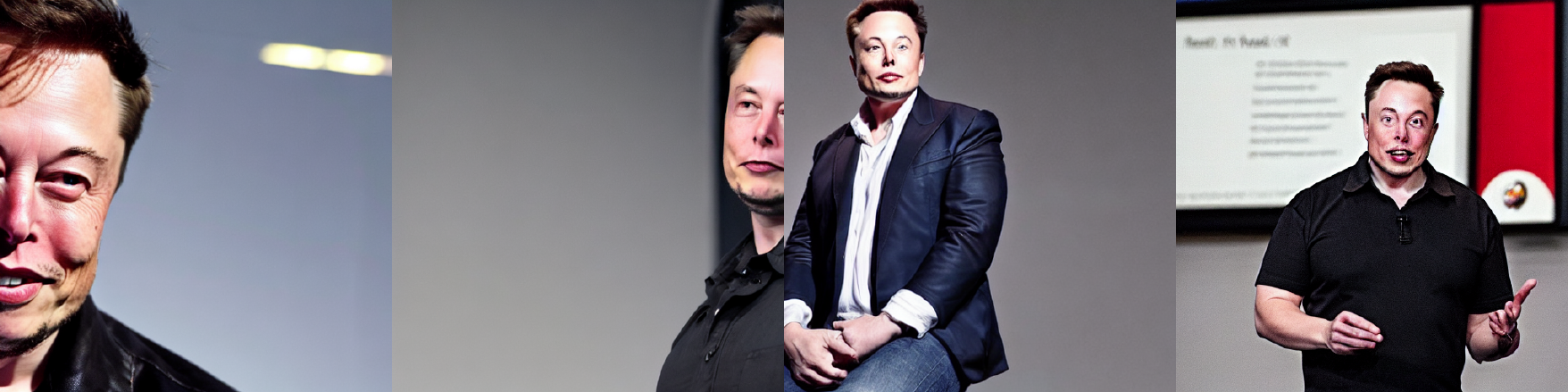}} \\ \hline % Adjusted image width
    \textbf{Topic Words} & \small \texttt{\{Photography and image quality\}, \{Public figures and entrepreneurship\},\{Technology and innovation\},\{Space exploration\}, \{Electric vehicles and sustainable energy\}, \{Influential personalities in modern industry\}, \{Media and public presence\}
    } 
    \\ \hline
    \(F_T\) & \small \texttt{\{Elon Musk\}, \{High Quality\}, \{Image\}} 
    \\ \hline
    \(F_I\) & \small \texttt{\{The individual depicted\}, \{Business attire\}, \{Casual attire\}, \{Presentation setting\}, \{Gesturing\} } \\ \hline
    \end{tabular}
    \end{adjustbox}
\end{table*}

\section{Defining Text-To-Image Model Memorization}

In this section, we explore various forms of memorization in text-to-image models. Initially, we provide a general definition of memorization within these models, followed by a categorization of different types of memorization in various scenarios.

\paragraph{Definition 1 (Memorization)} \textit{We define that model \( M \) memorizes the image feature \( f_i \) conditioned on text feature \( f_t \) if it replicates this feature in at least \( \delta\% \) of generations using different random initializations, while conditioned on \( f_t \).}

Users anticipate that the model will generate diverse features across different generations, and it is reasonable to expect some feature repetition when using the same prompt. However, if this repetition exceeds a certain threshold (\(\delta\)), there is cause for suspicion that the feature may have been memorized during training. This threshold might vary depending on the application and scenario.

Memorization has traditionally been viewed as a precursor to privacy breaches in machine learning models, prompting the development of various strategies to curb its effects without assessing their impact on utility and performance. However, the advent of generative models necessitates a reevaluation of this perspective. Is memorization always unintended? In some instances, users might expect the model to accurately generate the likeness of a famous character, especially if such characters are underrepresented in the training data, necessitating memorization. This leads us to propose new definitions for memorization that consider the scenario and user expectations.

\paragraph{Definition 2 (Explicit Intended Memorization \((M_E)\))} 
\textit{Explicit Intended Memorization (\(M_E\)) refers to a type of memorization that encompasses image features \(f_i\) which users expect the model to consciously and deliberately memorize and are present in the user's prompt. This typically includes features that are widely recognized and anticipated when using prompt \(p\).}

Now, we attempt to define \(M_E\) in mathematical terms. First, let \(F_T\) denote the set of text features of the given prompt, and let \(F_I\) represent the union set of image features from all generated images \(\{I_1, I_2, \ldots, I_N\}\) using different random initializations. Also, let \(F'_{I}\) denote the union of the image features of all training images, represented as \( \bigcup_{j=1}^{M} F_{I'_j} \), where \( F_{I'_j} \) is the set of image features for each training image \( I'_j \). Additionally, assume that \(M_E(F_T, F_I)\) is the set of explicit intended memorized features. The set \(M_E(F_T, F_I)\) possesses the following properties:

\begin{itemize}
    \item If \( f_i \in M_E(F_T, F_I) \implies f_i \in F'_I \).
    \item If \( f_i \in M_E(F_T, F_I) \implies \frac{|\{ j : f_i \in F_{I_j}\}|}{N} \geq \frac{\delta}{100} \).
    \item If \( f_i \in M_E(F_T, F_I) \implies f_i \in F_T \cap F_{I_i} \) for any \( I_i \in \{I_1, I_2, \ldots, I_N\} \).
\end{itemize}

Next, we turn our attention to defining the second type of memorization, which presents a different aspect of feature replication in generative models.

\paragraph{Definition 3 (Implicit Intended Memorization \((M_I)\)):} 
\textit{Implicit Intended Memorization \((M_I)\) refers to a type of memorization encompassing image features that the model has indirectly and conceptually memorized, which users might anticipate the model to exhibit, even though they are not explicitly stated in the prompt \(p\).}

Similar to the previous definition, we assume that \(M_I(F_T, F_I)\) represents the set of implicitly intended memorized features. Besides, let us assume that \( C_1, C_2, \ldots, C_k \) are the sets of features corresponding to the related concepts or topics associated with the prompt. With this premise, the set \( M_I(F_T, F_I) \), indicative of Implicit Intended Memorization, possesses the following properties:

\begin{itemize}
    \item If \( f_i \in M_I(F_T, F_I) \implies f_i \in F'_I \).
    \item If \( f_i \in M_I(F_T, F_I) \implies \frac{|\{ j : f_i \in (F_{I_j} - F_T)\}|}{N} \geq \frac{\delta}{100} \).
    \item If \( f_i \in M_I(F_T, F_I) \implies f_i \in \bigcup_{i=1}^{k} C_i \).

\end{itemize}

Having delineated both Explicit and Implicit Intended Memorization, we now turn our focus to the third and final category of memorization.

\paragraph{Definition 4 (Unintended Memorization \( M_U \)):} 
\textit{Unintended Memorization, denoted as \( M_U \), refers to features that the model has inadvertently memorized and that users typically do not anticipate. These features are neither present in the input text prompt nor do they belong to the related concepts or topics, yet they appear in at least \( \delta\% \) of the resulting generated images.}

Like other types of memorization, we define \( M_U(F_T,F_I) \) to be the set of unintended memorized features. We now proceed to detail the mathematical properties corresponding to this type of memorization:

\begin{itemize}
    \item If \( f_i \in M_U(F_T, F_I) \implies f_i \in F'_I \).
    \item If \( f_i \in M_U(F_T, F_I) \implies \frac{|\{ j : f_i \in (F_{I_j} - F_T)\}|}{N} \geq \frac{\delta}{100} \).
    \item If \( f_i \in M_I(F_T, F_I) \implies f_i \notin \bigcup_{i=1}^{k} C_i \).

\end{itemize}

With these definitions, we cover most scenarios of text-to-image model use, taking into account both the users' expectations and the important aspects of privacy and performance. Note that unintended memorization could include sensitive individual information. We will now proceed to discuss some details of these definitions.

\begin{figure}[ht] % Positioning options: h=here, t=top, b=bottom, p=separate page, !=override, you can use them in any combination, like [htbp]
    \centering
    \includegraphics[width=6cm, height=6cm]{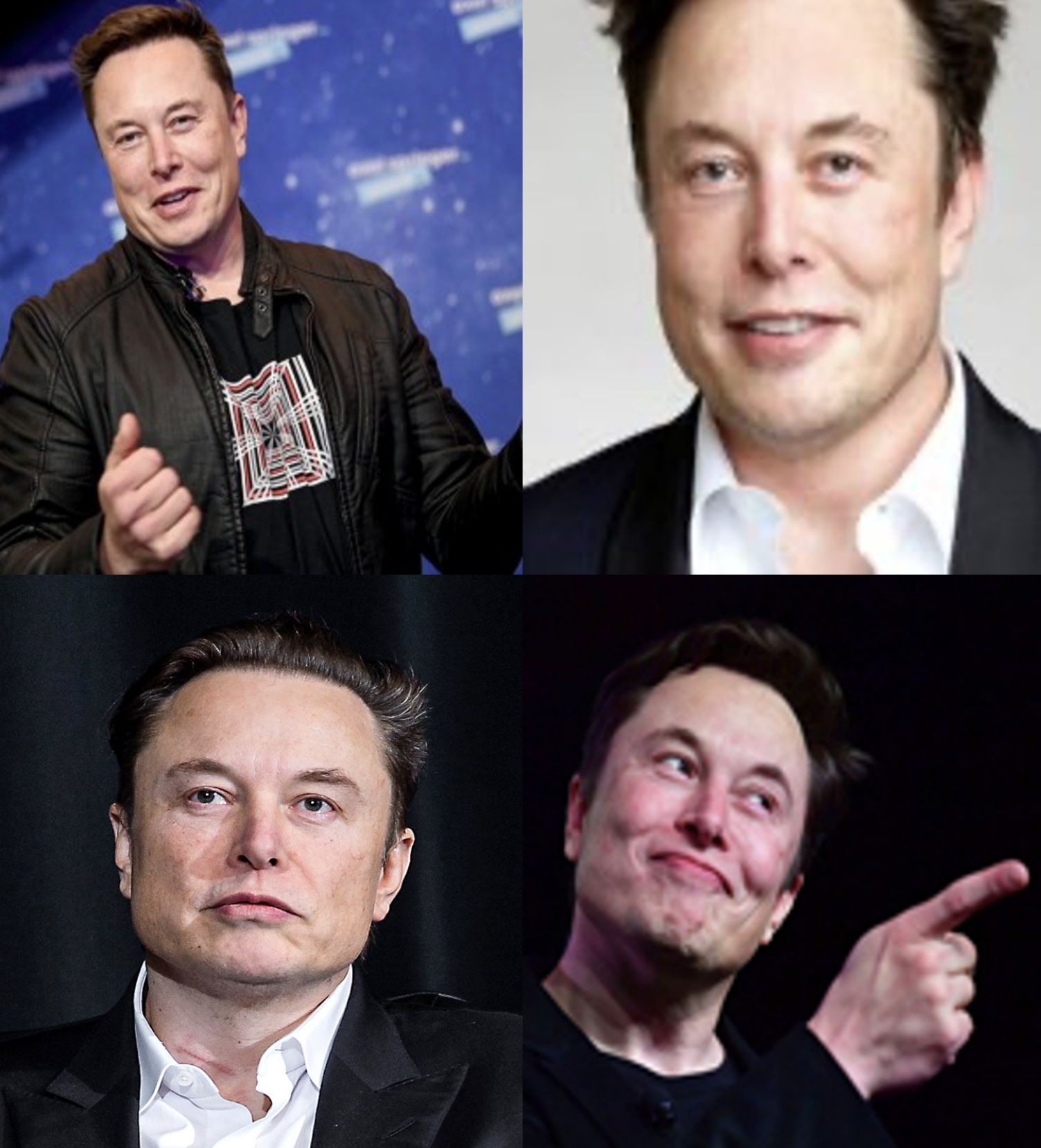}
    \caption{Training images from the LAION-5B dataset whose corresponding captions contain the name "Elon Musk".}
    \label{fig:elon_grid}
\end{figure}

\paragraph{Text and Image Features} In the context of the memorization types we have outlined, the terms ‘text features’ and ‘image features’ may carry diverse interpretations. For the purposes of this paper, these terms predominantly refer to the semantic features inherent in both modalities. Text features might include themes, entities, and emotions, while image features might encompass recognizable objects and activities. Thus, ‘text and image features’ within our defined framework specifically pertain to these semantic elements that bridge textual description and visual representation. Additionally, we use the term ‘concept’ or ‘topic’ of the text to refer to the main subjects, ideas, and entities that the text discusses or references.

Determining how to extract the semantic features of text and images is a pivotal step in understanding memorization within text-to-image models. This extraction process can be conducted manually by human annotators or automatically by leveraging deep learning models specialized in text and image processing. Notably, GPT-4 demonstrates remarkable proficiency in identifying such features from both text and images. Table 8 presents an example that showcases the efficacy of this model in feature extraction.

\begin{table*}[ht]
\centering
\caption{Example of Implicit Intended Memorization \(M_I\) based on the images created using pre-trained Stable Diffusion.}
    \label{tab:M_I_orig}
    \begin{adjustbox}{width=\textwidth,center}
    \begin{tabular}{>{\centering\arraybackslash}m{0.5\columnwidth} >{\arraybackslash}m{1.3\columnwidth}}
    \hline
    \textbf{Memorization Type} & \(M_I\) \\ \hline
    \textbf{Input Prompt} & \small \texttt{High resolution photo of Istanbul} \\ \hline
    \textbf{Generation} & {\includegraphics[width=9cm]{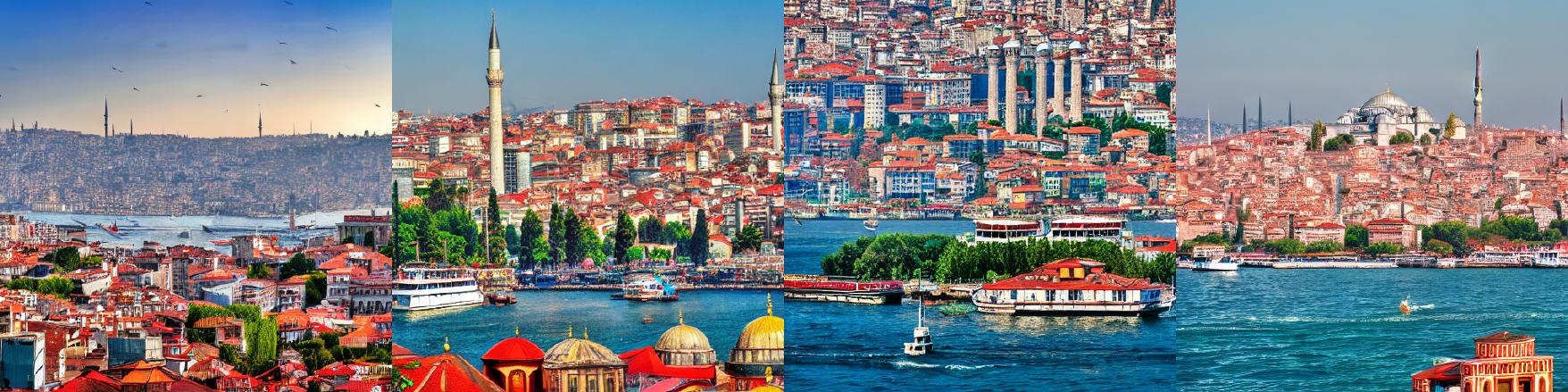}} \\ \hline
    \textbf{Topic Words} & \small \small \texttt{\{Photography and Imaging Technology\}, \{Istanbul's Landmarks and Attractions\}, \{Travel and Tourism in Istanbul\}, \{Urban Landscape and Architecture\}, \{Cultural Significance of Istanbul\}, \{Digital Media and Art\}, \{Geography and History of Istanbul\}} 
    \\ \hline
    
    \(F_T\) & \small \texttt{\{High resolution\}, \{Photo\}, \{Istanbul\}} \\ \hline
    
    \(F_I\) & \small \small \texttt{\{Body of water, likely the Bosphorus\}, \{Mosques with minarets\}, \{Dense urban housing\}, \{Boats and ferries\}, \{Natural landscape, including hills\}, \{Clear sky\},\{Trees and greenery\}} \\ \hline
    \end{tabular}
    \end{adjustbox}
\end{table*}

\begin{table*}[ht]
\centering
\caption{Example of Implicit Intended Memorization \(M_I\) based on the images created using pre-trained Stable Diffusion.}
    \label{tab:M_U_orig}
    \begin{adjustbox}{width=\textwidth,center}
    \begin{tabular}{>{\centering\arraybackslash}m{0.5\columnwidth} >{\arraybackslash}m{1.3\columnwidth}}
    \hline
    \textbf{Memorization Type} & \(M_U\) \\ \hline
    \textbf{Input Prompt} & \small \texttt{When a man came up with a brilliant idea} \\ \hline
    \textbf{Generation} & {\includegraphics[width=9cm]{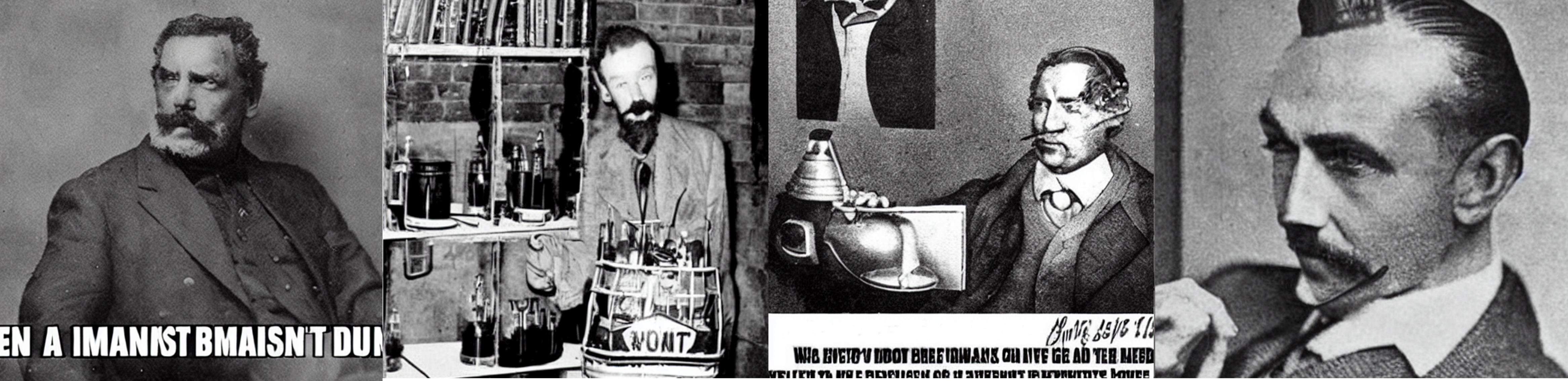}} \\ \hline
    \textbf{Topic Words} & \small \texttt{\{Creativity\}, \{Innovation\},  \{Invention\}, \{Problem-solving\}, \{Inspiration\}, \{Intellectual breakthroughs\}, \{Entrepreneurship\}}
    \\ \hline
    \(F_T\) & \small \texttt{\{Emergence of a new concept or invention\}, \{The involvement of a person\}, \{Moment of realization or discovery\}, \{The quality of idea\}}  
    \\ \hline
    \(F_I\) & \small \texttt{\{Portraits of men\}, \{Beards\}, \{Vintage attire\}, \{Scientific equipment or inventions\}, \{Text or writing accompanying the images\}, \{Pipe smoking\}} \\ \hline
    \end{tabular}
    \end{adjustbox}
\end{table*}

\section{Experimental Evaluation}

\begin{figure}[ht] % Positioning options: h=here, t=top, b=bottom, p=separate page, !=override, you can use them in any combination, like [htbp]
    \centering
    \includegraphics[width=\linewidth, height=5.7cm]{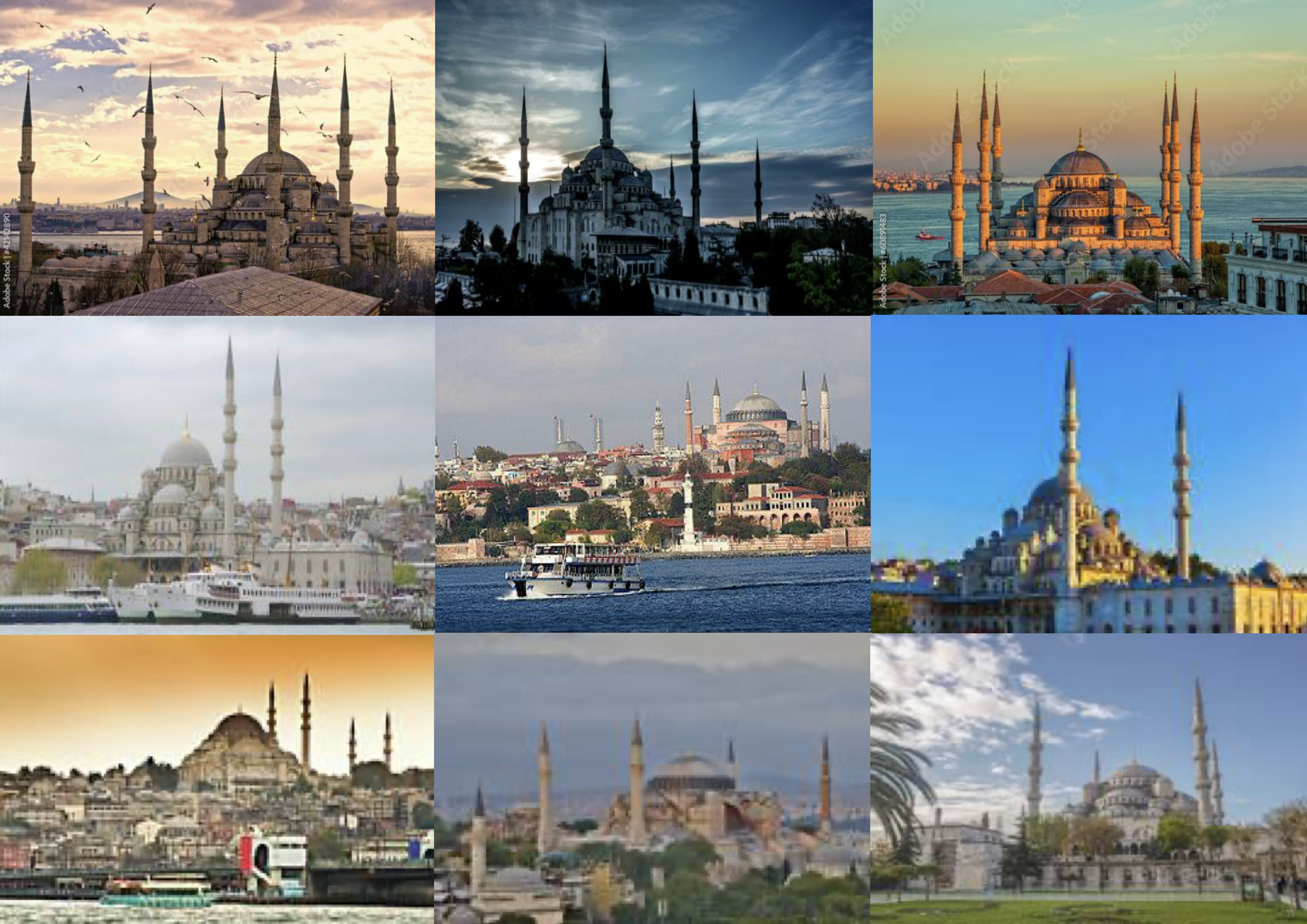}
    \caption{Training images from the LAION-5B dataset whose corresponding captions contain the word "Istanbul".}
    \label{fig:istanbul}
\end{figure}
\raggedbottom

In this section, we present a series of examples that provide a clearer illustration of the distinct characteristics inherent to each type of memorization defined previously, investigating these phenomena across both pre-trained and fine-tuned Stable Diffusion model. These examples are selected to shed light on how the model behaves in various scenarios, particularly focusing on the manifestation of memorization in the generated outputs.

\paragraph{Experimental Settings}
Our experiment employs the pre-trained Stable Diffusion model, initially trained on the LAION-5B dataset~\cite{schuhmann2022laion}, as well as the version fine-tuned on the Midjourney dataset for our experiment. In our approach, we employ the CompVis version 1-4 of the Stable Diffusion as our base model, initializing it with a randomly selected seed. During the fine-tuning stage, we train this model using the Midjourney dataset, which consists of 60,000 prompts and corresponding generated image pairs from the Midjourney API. Fine-tuning is conducted with a batch size of one, generating images with dimensions of 512 by 512 pixels, and we cap the training process at a maximum of 50,000 steps. 

\raggedbottom

\subsection{Explicit Intended Memorization (\(M_E\))}
The concept of explicit intended memorization (\(M_E\)) refers to features that the model is designed to memorize deliberately. Both Table~\ref{tab:M_E_orig} and Table~\ref{tab:M_E_mj} from Appendix highlight that the model is expected to produce an image of a renowned individual, in this case, “Elon Musk”, based on the input prompt. Observations suggest that the model adeptly learns to generate the specific figure. Moreover, the output from the fine-tuned model appears to mimic the distinctive image style from the Midjourney dataset. Examples of such generated images are provided in Fig.~\ref{fig:elon_grid}. Additionally, GPT-4 excels in extracting relevant topics and concepts from the provided input prompt. This capability significantly aids in assessing whether the generated images align particularly well with the implicit intended memorization. 

\begin{comment}
    The CLIP \cite{radford2021learning} Similarity Score quantifies the average cosine similarity of 100 CLIP image embeddings generated using the same input prompt, compared to the embedding of the top matching image from the dataset. Furthermore, each generated image corresponds to a matching training image, with the top match representing the training image that appears most frequently among the 100 generations. 
\end{comment}

%\input{AuthorKit24/AnonymousSubmission/LaTeX/sec/istanbul}

\subsection{Implicit Intended Memorization (\(M_I\))}
Implicit intended memorization (\(M_I\)) refers to features that the model has memorized conceptually, which are expected by users, but not explicitly stated in the prompt. For example, Table~\ref{tab:M_I_orig} shows that the term “Istanbul” in the prompt often results in images featuring the “Hagia Sophia Mosque”, despite it not being explicitly mentioned. This indicates the model's conceptual association of the location with the landmark, as reflected in its training data. 

\begin{figure}[ht] % Positioning options: h=here, t=top, b=bottom, p=separate page, !=override, you can use them in any combination, like [htbp]
    \centering
    \includegraphics[width=\linewidth, height=5.7cm]{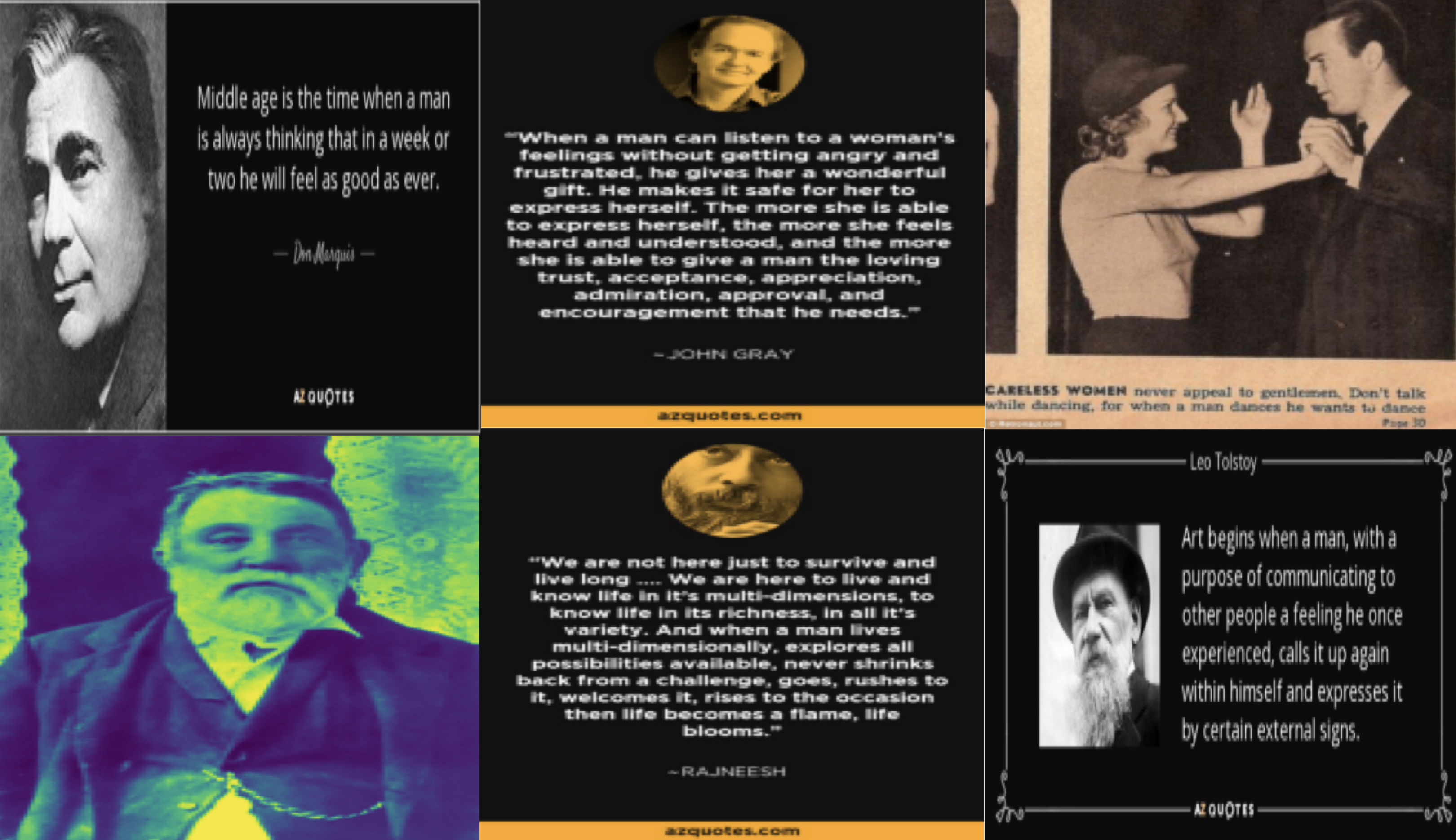}
    \caption{Training images from the LAION-5B dataset whose corresponding captions contain the phrase "When a man".}
    \label{fig:when}
\end{figure}

Table~\ref{tab:M_I_mj} reveals a consistent emergence of mosques with minarets in image generations, irrespective of their explicit mention in the captions. This pattern aligns with the expectation that iconic landmarks, such as the Hagia Sophia Mosque---an integral part of Istanbul's Landmarks and Attractions---would be featured in the model's output based on the relevant topics and concepts derived from the captions. The anticipation of these landmarks' appearance is based on the analysis of the training dataset. Filtering for instances where ``Istanbul'' is present, approximately 23\% of the images exhibit mosque features, suggesting that the model has learned to associate the city with this architectural element. Hence, there is a tendency for generated images to replicate mosques when ``Istanbul'' is referenced in the captions. Examples of such training images are displayed in Fig.~\ref{fig:istanbul}.

\subsection{Unintended Memorization (\(M_U\))}
Unintended memorization (\(M_U\)) occurs when the model unexpectedly memorizes and includes features in its output that users may not anticipate. The pre-trained model from Table~\ref{tab:M_U_orig} produces a variety of images, predominantly featuring different male figures. Unexpected features such as a “beard,” the image being “black and white,” or the presence of “text within the image,” derived from the top matches in the LAION-5B dataset, appear without user intent, exemplifying unintended memorization. Additionally, since the model recreates specific facial features from a non-public figure in the training dataset, this scenario may raise concerns of sensitive information exposure or privacy leakage in Stable Diffusion models.  

Our experimentation with varying the terms used in the prompts, as detailed in Table~\ref{tab:M_U_orig}, revealed that the words "when" and "man" significantly influence the repetition of certain features. Upon examining the training dataset, we observed that approximately 5\% of the samples containing the phrase "when a man" were rendered in black and white, and they frequently included text within the image. This suggests a correlation between the specific language of the prompt and the characteristics of the generated images, underscoring the impact of dataset composition on model behavior. Representative examples of these training images are illustrated in Fig.~\ref{fig:when}.

\begin{figure}[htbp] % Positioning options: h=here, t=top, b=bottom, p=separate page, !=override, you can use them in any combination, like [htbp]
    \centering
    \includegraphics[width=5cm, height=8.5cm]{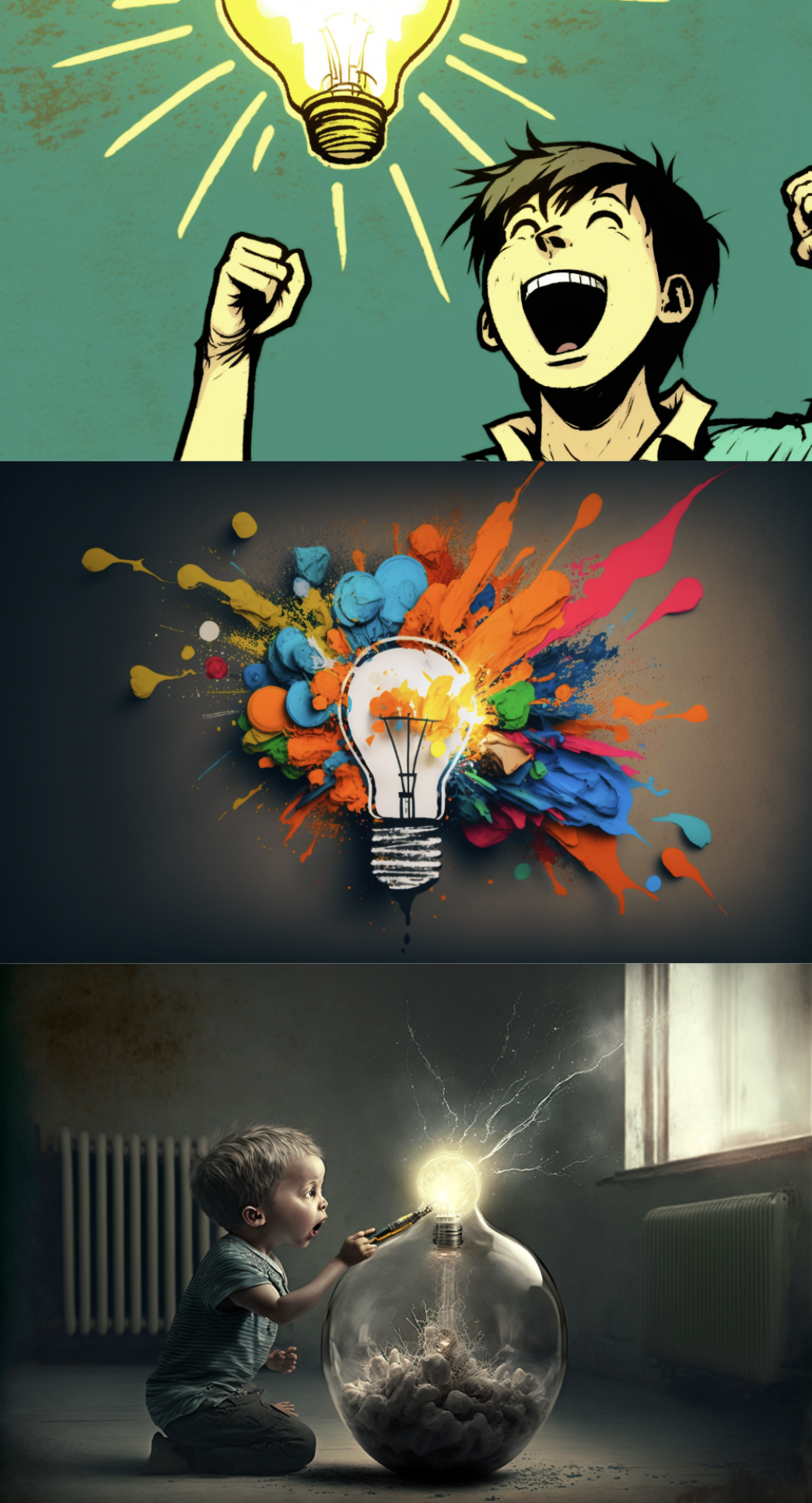}
    \caption{Presence of light bulb in the training images from the Midjourney dataset that contain the word "idea" in the caption.}
    \label{fig:idea}
\end{figure}

Additionally, the fine-tuned model's output, which prominently features a light bulb, omits the man mentioned in the prompt. As shown in Table~\ref{tab:M_U_mj}, it is likely that the term “idea” in the text prompt triggered the inclusion of the light bulb, a detail present in the Midjourney dataset, resulting in an unexpected element in the image from the user's perspective. Around 5\% of fine-tuning samples whose text contains the word "idea" feature a light bulb in their corresponding images, as shown in Fig.~\ref{fig:idea}. Although only a small fraction of the training samples, this is notable within a dataset of 60,000 samples. Also, since the correlation between “black and white” as an image feature with the terms "when a man" in the pre-training dataset is not strong, the model, after fine-tuning, unintentionally memorized the “light bulb” when conditioned on text containing the word “idea”.

% \subsection{Perplexed Unintended Memorization}

\begin{comment}
\subsection{Challenges and Future Works}
For the advancement of this research, we aim to undertake a broader experimental approach by collecting a more extensive ensemble of examples for each memorization category to quantitatively assess memorization frequencies. Additionally, we encourage future investigators to develop novel membership inference attacks based from our listed memorization instances, with a goal of developing innovative training methodologies or defensive strategies to enhance the robustness of text-to-image models. 
\end{comment}
\section{Challenges and Future Work}

Evaluating memorization using our definitions, especially with generated images from given prompts, involves complex tasks. Key among these is finding a model like GPT-4, which excels in feature extraction but is costly for large-scale use. Efficiently determining the occurrence of repeated features in extensive training datasets is another challenge that requires future research focus.

Current strategies to mitigate memorization, such as deduplication and differential privacy, often diminish generative model performance. A vital research direction is the development of more balanced mitigation methods that ensure utility. This underscores the need for comprehensive memorization definitions that meet diverse goals. Such efforts are crucial for advancing generative models that perform effectively while safeguarding privacy through careful memorization management.

\section{Conclusion}

While memorization in text-to-image models has been partially explored, a comprehensive understanding that balances privacy with utility remains elusive. In this paper, we offer a general definition of memorization and delineated three categories tailored to user expectations and various application scenarios. By providing concrete examples, we bring more clarity to these categories. Our contributions serve as a foundation for future endeavors to refine these definitions and develop mitigation strategies that preserve the functional integrity of generative models.

\bibliography{main}

\section{Appendix}
% \begin{figure}[htbp] % Positioning options: h=here, t=top, b=bottom, p=separate page, !=override, you can use them in any combination, like [htbp]
%     \centering
%     \includegraphics[width=\linewidth]{AuthorKit24/AnonymousSubmission/images/gpt4_caption.pdf}
%     \caption{Images produced by Stable Diffusion \cite{rombach2022high}, finetuned using Midjourney \cite{midjourney2022}, with an input caption as "campus life in autumn".}
%     \label{fig:autumn}
% \end{figure}

% \begin{figure}[htbp] % Positioning options: h=here, t=top, b=bottom, p=separate page, !=override, you can use them in any combination, like [htbp]
%     \centering
%     \includegraphics[width=\linewidth]{AuthorKit24/AnonymousSubmission/images/astronaut's wife.pdf}
%     \caption{Images produced by Stable Diffusion \cite{rombach2022high}, fine-tuned using Midjourney \cite{midjourney2022}, with an input caption as "\textit{astronaut's wife movie poster}".}
%     \label{fig:astronaut}
% \end{figure}

% \begin{figure}[htbp] % Positioning options: h=here, t=top, b=bottom, p=separate page, !=override, you can use them in any combination, like [htbp]
%     \centering
%     \includegraphics[width=\linewidth]{AuthorKit24/AnonymousSubmission/images/new_idea_mj_gen.pdf}
%     \caption{Images produced by Stable Diffusion \cite{rombach2022high}, fine-tuned using Midjourney \cite{midjourney2022}, with an input caption as "\textit{birth of a new idea –v 4 –q 2 –ar 3:2}".}
%     \label{fig:new_idea_gen}
% \end{figure}

\begin{table*}[ht]
\centering
\caption{Example of Explicit Intended Memorization \(M_E\) based on the images created using Midjourney's fine-tuned version of Stable Diffusion.}
    \label{tab:M_E_mj}
    \begin{adjustbox}{width=\textwidth,center}
    \begin{tabular}{>{\centering\arraybackslash}m{0.5\columnwidth} >{\arraybackslash}m{1.3\columnwidth}}
    \hline
    \textbf{Memorization Type} & \(M_E\) \\ \hline
    \textbf{Input Prompt} & \small \texttt{High quality image of Elon Musk} \\ \hline
    \textbf{Generation} & {\includegraphics[width=9cm]{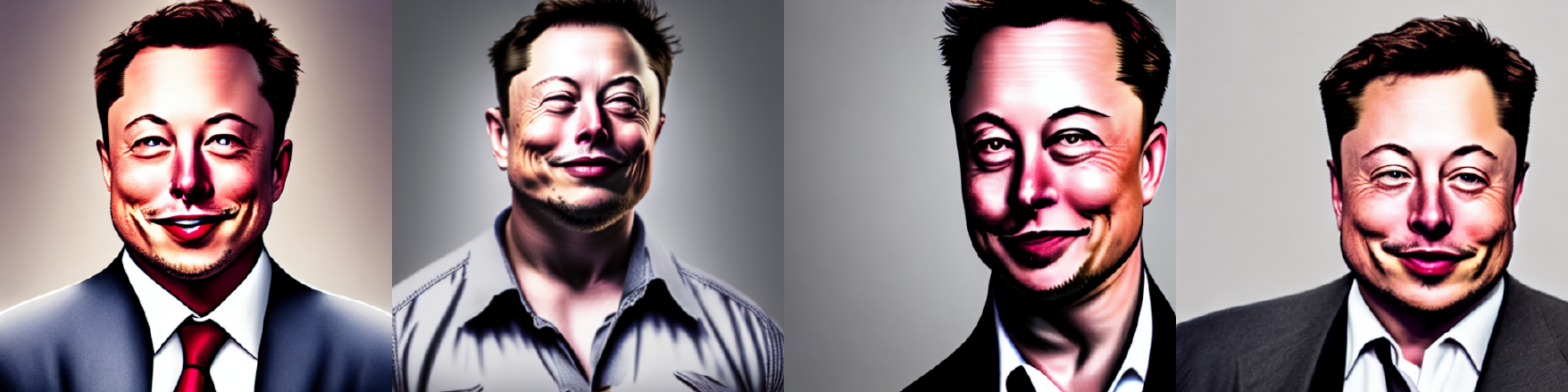}} \\ \hline
    \textbf{Topic Words} & \small \texttt{\{Photography and image quality\}, \{Public figures and entrepreneurship\}, \{Technology and innovation\}, \{Space exploration\}, \{Electric vehicles and sustainable energy\}, \{Influential personalities in modern industry\}, \{Media and public presence\}} 
    \\ \hline
    \(F_T\) & \small \texttt{\{Elon Musk\}, \{High Quality\}, \{Image\}} 
    \\ \hline
    \(F_I\) & \small \texttt{\{Depictions of a smiling man\}, \{Wearing a suit\}, \{Cartoonish\}, \{High contrast with brush strokes\}, \{Realistic\}, \{Mix of realism with saturation\}, \{Portrayal of headshots\}, \{Grinning\}, \{Smiling with closed and open mouth\}, \{Neutral\}, \{Well-groomed\}} \\ \hline
    \end{tabular}
    \end{adjustbox}
\end{table*}

\begin{table*}[ht]
\centering
\caption{Example of Implicit Intended Memorization \(M_I\) based on the images created using Midjourney's fine-tuned version of Stable Diffusion.}
    \label{tab:M_I_mj}
    \begin{adjustbox}{width=\textwidth,center}
    \begin{tabular}{>{\centering\arraybackslash}m{0.5\columnwidth} >{\arraybackslash}m{1.3\columnwidth}}
    \hline
    \textbf{Memorization Type} & \(M_I\) \\ \hline
    \textbf{Input Prompt} & \small \texttt{High resolution photo of Istanbul} \\ \hline
    \textbf{Generation} & {\includegraphics[width=9cm]{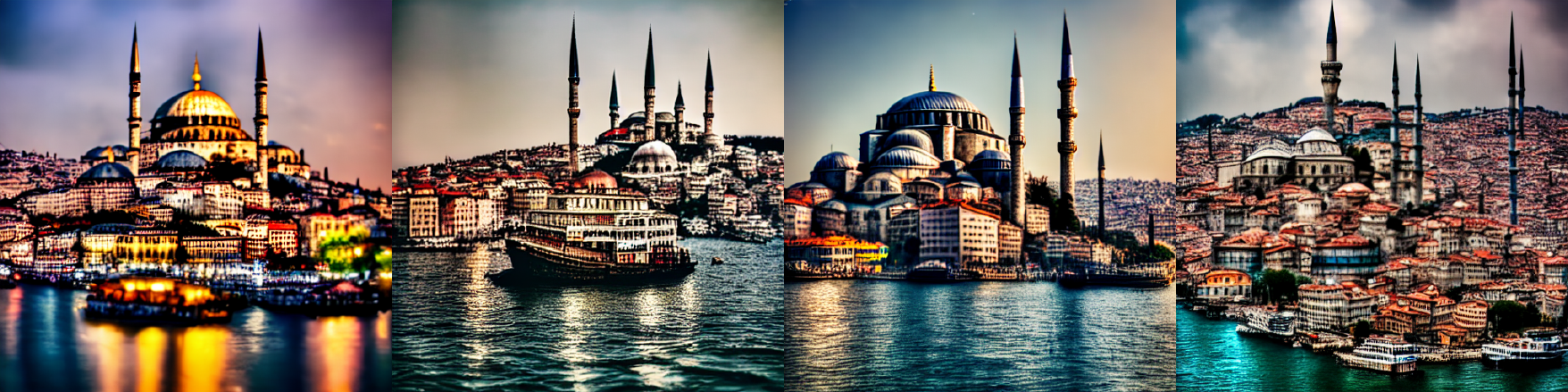}} \\ \hline
    \textbf{Topic Words} & \small \texttt{\{Photography and Imaging Technology\}, \{Istanbul's Landmarks and Attractions\}, \{Travel and Tourism in Istanbul\}, \{Urban Landscape and Architecture\}, \{Cultural Significance of Istanbul\}, \{Digital Media and Art\}, \{Geography and History of Istanbul\}} \\ \hline
    
    \(F_T\) & \small \texttt{\{High resolution\}, \{Photo\}, \{Istanbul\}}
    \\ \hline
    
    \(F_I\) & \small \texttt{\{Mosques with prominent domes and minarets\}, \{Waterfront with marine traffic\}, \{Urban landscape with dense buildings\},  \{Hillsides as part of the city's topography\}, \{The Bosphorus, which is the body of water featured\}, \{The evening or night-time lighting, with city lights and reflections on water\}, \{Skyline with a mix of modern and historic architecture\}} \\ \hline
    \end{tabular}
    \end{adjustbox}
\end{table*}

\begin{table*}[ht]
\centering
\caption{Example of Unintended Memorization \(M_U\) based on the images created using Midjourney's fine-tuned version of Stable Diffusion.}
    \label{tab:M_U_mj}
    \begin{adjustbox}{width=\textwidth,center}
    \begin{tabular}{>{\centering\arraybackslash}m{0.5\columnwidth} >{\arraybackslash}m{1.3\columnwidth}}
    \hline
    \textbf{Memorization Type} & \(M_U\) \\ \hline
    \textbf{Input Prompt} & \small \texttt{When a man came up with a brilliant idea} \\ \hline
    \textbf{Generation} & {\includegraphics[width=9cm]{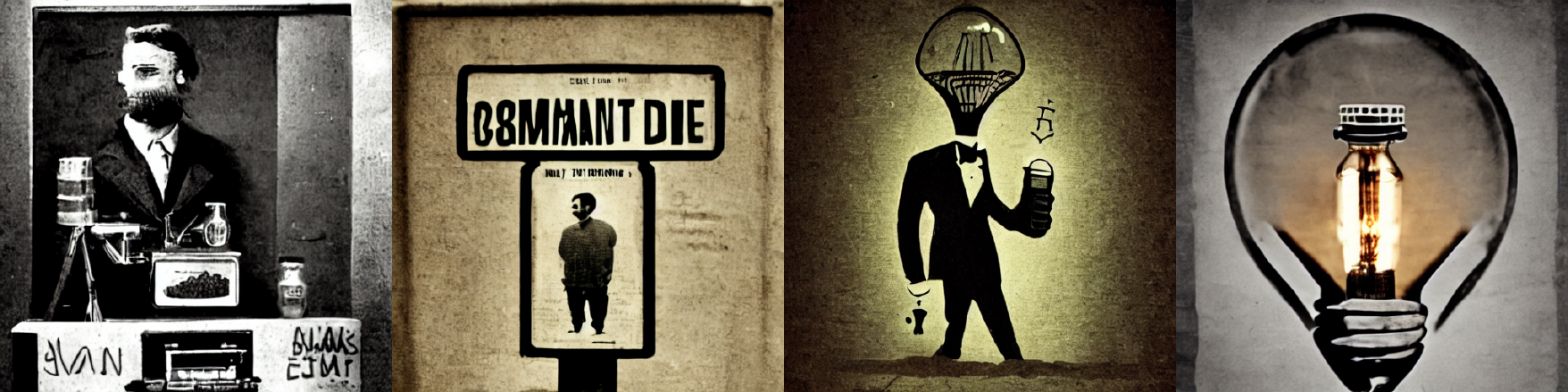}} \\ \hline
    \textbf{Topic Words} & \small \texttt{\{Creativity\}, \{Innovation\}, \{Invention\}, \{Problem-solving\}, \{Inspiration\}, \{Intellectual breakthroughs\}, \{Entrepreneurship\}
    }
    \\ \hline
    \(F_T\) & \small \texttt{\{Emergence of a new concept or invention\}, \{The involvement of a person\}, \{Moment of realization or discovery\}, \{The quality of idea\}} 
    \\ \hline
    \(F_I\) & \small \texttt{\{Light bulbs\}, \{Men\}, \{Scientific or experimental equipment\}, \{Text or writing\}, \{Dark suit attire\}, \{Silhouette form\}} \\ \hline
    \end{tabular}
    \end{adjustbox}
\end{table*}

\end{document}